\documentclass{ADAPTCONF2025}

\interspeechcameraready 
\usepackage{multirow}
\usepackage{caption}
\usepackage{subcaption}

\title{Pre-Hoc Predictions in AutoML: Leveraging LLMs to Enhance Model Selection and Benchmarking for Tabular datasets}
\name{Yannis Belkhiter$^1$$^2$, Seshu Tirupathi$^1$, Giulio Zizzo$^1$, Sachin Sharma$^3$, John D. Kelleher$^2$}
\address{
  $^1$IBM Research Europe,
  $^2$Trinity College Dublin,
  $^3$Technological University Dublin
  }
\email{yannis.belkhiter@ibm.com, seshutir@ie.ibm.com, giulio.zizzo2@ibm.com, sachin.sharma@tudublin.ie, 
john.kelleher@tcd.ie}

\begin{document}

\maketitle
The field of AutoML has made remarkable progress in post-hoc model selection, with libraries capable of automatically identifying the most performing models for a given dataset. Nevertheless, these methods often rely on exhaustive hyperparameter searches, where methods automatically train and test different types of models on the target dataset. Contrastingly, pre-hoc prediction emerges as a promising alternative, capable of bypassing exhaustive search through intelligent pre-selection of models. Despite its potential, pre-hoc prediction remains under-explored in the literature. This paper explores the intersection of AutoML and pre-hoc model selection by leveraging traditional models and Large Language Model (LLM) agents to reduce the search space of AutoML libraries. By relying on dataset descriptions and statistical information, we reduce the AutoML search space. Our methodology is applied to the AWS AutoGluon portfolio dataset, a state-of-the-art AutoML benchmark containing 175 tabular classification datasets available on OpenML. The proposed approach offers a shift in AutoML workflows, significantly reducing computational overhead, while still selecting the best model for the given dataset.

\section{Introduction}

Post-hoc AutoML model selection refers to the process of finding the most performant model for a given dataset. Over the past few years, various AutoML libraries have been published such as AutoGluon \cite{erickson2020autogluontabularrobustaccurateautoml}, H2O \cite{LeDell2020H2OAS}, and TPOT \cite{olson2016evaluationtreebasedpipelineoptimization}. However, even though post-hoc AutoML model selection is performing well, it needs significant computation to run these searches. The computational demands of these frameworks remain a significant barrier, especially for users with limited access to computing infrastructure. 

To address these challenges, pre-hoc AutoML has emerged as a promising alternative \cite{santu2021automldatebeyondchallenges,tornede2024automlagelargelanguage}, aiming to identify the most suitable model or pipeline for a given dataset \emph{before} running extensive computations. Nevertheless, to the best of our knowledge, no previous work has investigated the use of Large Language Models (LLMs) as a pre-hoc model selection alternative in real-world scenarios, such as those involving OpenML datasets, and directly compared the results to a top-performing post-hoc AutoML library. This paper helps to address key gaps in pre-hoc AutoML research by proposing a novel framework. Our contributions are as follows:

\begin{itemize}[leftmargin=*]
    \item \textbf{Integration of diverse dataset information:} We tested the integration of structured statistical information about datasets, with textual dataset descriptions, and retrieval-augmented generation (RAG) to enable efficient and accurate pre-hoc model selection.

    \item \textbf{AutoML agent:} By leveraging LLMs with enhanced context, we developed a pipeline for solving tabular AutoML problems. This approach predicts suitable models, and also enables models to provide reasoning justification of their selection, contributing to explainability.

    %\item \textbf{OpenML dataset:} To ensure that our experiences are rooted in a real-world environment, we ran our experiments on datasets from the OpenML repository \cite{openml}.
\end{itemize}

\section{Related work}

Several studies have explored pre-hoc AutoML. For instance, the Meta-Learning approach described in \cite{vanschoren2018metalearningsurvey} and \cite{Kotthoff2019} leverages knowledge from prior experiments to recommend models based on dataset properties. Similarly, \cite{rakotoarison2022learning,nápoles2022bestmodeldata} introduced techniques to predict which model will perform best, leveraging metadata and historical model performance to inform their decisions. On the intersection of LLMs with AutoML, recent works made significant progresses by building frameworks capable of enhancing LLM reasoning for specific tasks. \cite{zhang2024mlcopilotunleashingpowerlarge} proposed an LLM-based framework that leverages existing model performance on datasets to aid model selection and optimization. Likewise, \cite{liu2025largelanguagemodelagent} implemented a novel LLM-based framework specifically for hyperparameter optimization. 

\section{Pre-hoc pipeline}\label{sec:pre_hoc_pipeline}

The goal of this paper is to demonstrate that numerical or textual descriptions of a dataset can be leveraged to suggest a good performing model to the user. By relying on this supplementary condensed information, we can enhance the model selection process and automate the task of choosing an appropriate machine learning model based on the dataset at hand. We explored two different ways of enhancing the dataset information: numerical statistical information about a dataset, and the textual dataset description provided by OpenML \cite{openml}.

\subsection{Dataset metadata: Statistical information}

We use the term \emph{metadata} to refer to statistical information describing the given dataset \cite{vanschoren2018metalearningsurvey}. We compute statistical variables for each dataset contained in the portfolio in order to infer pre-hoc information about model performances. Inspired by \cite{nápoles2022bestmodeldata}, the selected statistical variables are listed in Table \ref{tab:stat_variables}.

\subsection{Textual dataset description}

As datasets are at the heart of Machine Learning, it is essential for each data provider to include a dataset card, containing information and description about the dataset. In our experimentation, each dataset included in the portfolio is provided by OpenML \cite{openml}. OpenML equips each dataset with the name and domain of the dataset, some information about the features, the type of task, and references about sources and authors.

\section{Experimental Setup}\label{sec:exp-setup}

\vspace{-0.1cm}

Our experimentation consists in testing different methods for pre-hoc model selection given a dataset. Two strategies are explored. First, the traditional Pre-Hoc Predictors (Pre-HP) methods, including ensemble methods or BERT encoding. Second, the LLM Pre-HPs, with different ways of enhancing their context to improve their prediction. In order to compare our different strategies, we use 175 datasets available on OpenML \cite{openml}. 

\vspace{-0.1cm}

\subsection{TabRepo: AWS AutoGluon portfolio}

\vspace{-0.1cm}

In the previous work published by \cite{erickson2020autogluontabularrobustaccurateautoml}, authors presented a major study on the efficiency of the \emph{AutoGluon} algorithm on a large number of open-source datasets available on OpenML \cite{openml}. It provided a large amount of information for each dataset, including runtimes and training/testing metrics of 11 different models. The datasets are divided in 3 folds, and performances are given for each fold. To conduct our experiments, we select the following configuration: \textbf{``D244\_F3\_C1530\_175''} \cite{autogluon_tabrepo}, including 175 classification and regression datasets and 4,590 model evaluations per dataset. Further details of the portfolio are described in Section \ref{sec:dataset-analysis}.

\vspace{-0.15cm}

\subsection{Available Tabular Models}

\vspace{-0.1cm}

We can observe that most Machine Learning (ML) libraries consistently implement the same state-of-the-art models. Table \ref{tab:model-families} in Section \ref{sec:available-tab-models} references the 11 models included by the AutoGluon AutoML library, along with their associated families and hyper-parameters. For each dataset, the TabRepo includes a ranking of each of these model based on their performance.

\vspace{-0.15cm}

\subsection{Traditional methods}

\vspace{-0.1cm}

Using what we refer as traditional methods, our first approach involves training traditional Pre-HP models using the most performant model as per TabRepo as the target label, based on either metadata or textual descriptions. Traditional Pre-HP models used in our experiments are outlined in Table \ref{tab:traditional-models} in Section \ref{sec:trad-pre-hp-models}. These models rely on simple ML methods, yet are effective and explainable.

\vspace{-0.15cm}

\subsection{Auto-ML agent}

\vspace{-0.1cm}

Based on metadata of datasets, and RAG content about available tabular models, our second method relies on LLMs. Our objective is to build an Auto-ML agent capable of accurately selecting a model from dataset information. To enhance the knowledge of LLM, we implemented a \emph{RAG agent} using \emph{Llama-70b}, to retrieve information about available models, but also additional insights on their usage from the documentation of each libraries. To test different configurations, we infer models using \emph{Zero-Shot} or \emph{Few-Shot}. In Few-Shot, we add one metadata example for each available model from its best performing label. We remove this information in Zero-Shot setting. We test Granite-3.1-8b \cite{granite3_8b_instruct}, Llama-3.1-8b \cite{llama3_8b_instruct}, and GPT-4o \cite{openai_gpt4o}.

\vspace{-0.15cm}

\subsection{Metrics}

\vspace{-0.1cm}

To rank performances of each approaches, we use two metrics:

\begin{itemize}[leftmargin=*, noitemsep]
    \item We group models by family based on their underlying characteristics (see Table \ref{tab:model-families} in Section \ref{sec:available-tab-models}). The first objective is to assess whether a Pre-HP technique is capable of selecting the correct ground-truth family of models.
    \item Secondly, we match the ground truth label with the one predicted by the Pre-HP methods.
\end{itemize}

\noindent For either metric, we compute the accuracy. To compare each approach, we include two baselines: (1) one assumes a random pick with a uniform probability for each model \footnote{To obtain Baseline (1) metrics, we average the score of 1,000 experiments on Ground-Truth labels}, and (2) a second always selects the most frequent label of the portfolio.

\vspace{-0.2cm}

\section{Evaluation}

\vspace{-0.1cm}

\subsection{Traditional Pre-HP methods}

\begin{table}[h]
\vspace{-0.4cm}
\centering
\footnotesize
\setlength{\tabcolsep}{2pt}
\renewcommand{\arraystretch}{1}
\begin{tabular}{@{}llcc@{}}
\toprule
\textbf{Source} & \textbf{Method} & \textbf{Family Accuracy} & \textbf{Model Accuracy} \\ \midrule
\multirow{2}{*}{-} 
                          & Baseline 1 (Random) & 0.2086 & 0.0920 \\ 
                          & Baseline 2 (Frequency) & 0.4228 & 0.2400 \\
                          \midrule
\multirow{3}{*}{Metadata} 
                          & Euclidean Distance & 0.4286 & 0.2857 \\
                          & KNN (n\_neigh=3) & 0.4571 & \textbf{0.3714} \\
                          & RFC (n\_tress=100) & 0.5143 & \textbf{0.3714} \\ \midrule
                          & TF-IDF encoding & 0.4444 & 0.2222 \\
Dataset                   & BERT encoding & 0.5000 & 0.1667 \\
Description               & RoBERTa encoding & \textbf{0.6111} & 0.2222 \\
                          & BERT Classifier & 0.4000 & 0.2000 \\ \bottomrule
\end{tabular}
\caption{\centering Performance of Traditional Pre-hoc Predictors}
\label{tab:experiment-results-traditional}
\vspace{-0.7cm}
\end{table}

\noindent Table \ref{tab:experiment-results-traditional} presents the performance of Traditional Pre-HP models over the AutoGluon portfolio. As each model required data for training, we split the portfolio using stratified sampling based on the target labels (80:20 train-test). First, we observe that most traditional models outperform the baselines for family and model selection, suggesting that both statistical information and dataset description contains relevant information for the prediction. Pre-HP models relying on metadata present good results for model accuracy, and Pre-HP methods using dataset description show promising results for Family selection, especially the RoBERTa model with a Family accuracy of 0.61. 

\vspace{-0.1cm}

\subsection{AutoML agent}

\begin{table}[h]
\vspace{-0.4cm}
\centering
\footnotesize
\setlength{\tabcolsep}{2pt} 
\renewcommand{\arraystretch}{1} 
\begin{tabular}{@{}llcc@{}}
\toprule
\textbf{Configuration} & \textbf{Model} & \textbf{Family Accuracy} & \textbf{Model Accuracy} \\ 
\midrule
Random Guess        & \textbf{Baseline 1}  & 0.2086 & 0.0920     \\
Most frequent label  & \textbf{Baseline 2} & 0.4228 & 0.2400 \\
\midrule
& & \multicolumn{2}{c}{\textbf{Zero-Shot}} \\ 
\midrule
\multirow{3}{*}{No RAG} 
             & \textbf{GPT-4o}        & 0.3828 & 0.1600 \\  
             & \textbf{Llama 3.1 8b}  & \textbf{0.3829} & \textbf{0.2000} \\ 
             & \textbf{Granite 3.1 8b} & 0.3600 & 0.1314 \\ 
\midrule
\multirow{3}{*}{With RAG} 
             & \textbf{GPT-4o}        & 0.3142 & 0.1028 \\  
             & \textbf{Llama 3.1 8b}  & 0.3828 & 0.1486 \\ 
             & \textbf{Granite 3.1 8b} & 0.3600 & 0.1600 \\ 
\midrule
& & \multicolumn{2}{c}{\textbf{Few-Shot}} \\ 
\midrule
\multirow{3}{*}{No RAG} & \textbf{GPT-4o}  & 0.2857 & 0.0742 \\  
                            & \textbf{Llama 3.1 8b} & 0.2457 & 0.0800 \\  
                            & \textbf{Granite 3.1 8b} & 0.3714 & 0.1542 \\ \midrule
\multirow{3}{*}{With RAG} & \textbf{GPT-4o}  & 0.3371 & 0.0685 \\  
                            & \textbf{Llama 3.1 8b}  & 0.2800 & 0.0571 \\  
                            & \textbf{Granite 3.1 8b} & 0.3657 & 0.1429 \\
\bottomrule
\end{tabular}
\caption{\centering Results for Pre-hoc LLM Predictors}
\label{tab:experiment-results-llm-agent}
\vspace{-0.7cm}
\end{table}

\noindent Table \ref{tab:experiment-results-llm-agent} provides the results of the AutoML agent for pre-hoc selection, where the entire portfolio serves as the test dataset. First, results are notably lower than traditional models, while still outperforming  Baseline 1. However, this is expected, as Traditional Pre-HP models have access to more knowledge from previous experiments, providing them with a significant advantage. Comparing the different results obtained, Llama in Zero Shot No Rag setting has best performance overall across metrics and Granite provided more consistency in predictions over the different settings. The addition of RAG only has an impact on the Few-Shot setting, significantly increasing the Family accuracy for GPT and Llama. Unexpectedly, while more Few-Shot examples could be added, Few-Shot configurations seem to not increase performances over Zero-Shot settings.

\vspace{-0.2cm}

\section{Conclusion}

\vspace{-0.1cm}

Our approaches demonstrate promising potential for improving pre-hoc model selection in AutoML. Leveraging traditional methods, we highlighted the value of dataset characterization as metadata and textual descriptions for inferring knowledge about model performance. Our experiments showed that traditional models could benefit from enriched information of datasets, while LLMs exhibit, with further refinement needed, reasoning capabilities on AutoML problems. Overall, our work paves the way for more efficient AutoML approaches, and promotes the value of open-source datasets with consistent textual and numerical information.

\newpage

\bibliographystyle{IEEEtran}
\bibliography{adaptconf2025}

\newpage

\appendix
\section{Appendix}

\subsection{Statistical variable composing the metadata}

\vspace{-0.2cm}

\begin{table}[ht]
\centering
\begin{tabular}{ll}
\hline
Number of samples          & Average skewness        \\
Number of features         & Average kurtosis        \\
Num. numerical features    & Average variance        \\
Num. categorical features  & Number of missing values\\
Class imbalance            & Target entropy          \\
Number of outliers         &                         \\
\hline
\end{tabular}
\caption{Selected Statistical Variables}
\label{tab:stat_variables}
\end{table}

\vspace{-0.8cm}

\subsection{Dataset analysis}\label{sec:dataset-analysis}

In this section, we present an overview of the dataset characteristics. Table \ref{tab:dataset_summary} includes statistics about the datasets composing the portfolio. 

\begin{table}[h]
\centering
\footnotesize
\begin{tabular}{lccc}
\toprule
\textbf{Classification type}                          &  & \textbf{Value} & \\ 
\midrule
Number of Datasets              & \multicolumn{3}{c}{175} \\ 
Binary Classification           & \multicolumn{3}{c}{95} \\ 
Multiclass Classification       & \multicolumn{3}{c}{60} \\ 
Regression     & \multicolumn{3}{c}{20} \\
\midrule
\textbf{Statistic}                          & \textbf{Min} & \textbf{Max}    & \textbf{Mean}      \\ 
\midrule
Number of Samples               & 100          & 53,940         & 5,486.60          \\ 
Number of Features              & 4            & 10,936         & 643.41            \\ 
Number of Missing Values        & 0            & 104,249        & 1,247.03          \\ 
Percentage of Missing Values    & 0.0          & 0.3279         & 0.01              \\ 
Number of Classes               & 2            & 28            & 4.10            \\ 
Average Skewness                & -5.0556      & 10,000.0       & 60.24             \\ 
Average Kurtosis                & -1.9993      & 10,000.0       & 206.39            \\ 
\bottomrule
\end{tabular}
\caption{Dataset Summary Statistics}
\label{tab:dataset_summary}
\end{table}

\vspace{-0.8cm}

\subsection{Available Tabular models}\label{sec:available-tab-models}

\vspace{-0.1cm}

Table \ref{tab:model-families} describes the available models in the \emph{AutoGluon} library. These models are further classified by Model Family. All the models mentioned in Table \ref{tab:model-families} are considered for the pre-hoc AutoML analysis in this paper. 

\begin{table}[h]
\centering
\footnotesize
\renewcommand{\arraystretch}{1.1}
\begin{tabular}{@{}ll@{}}
\toprule
\textbf{Model Family}      & \textbf{Model}            \\ \midrule
Boosting Methods           & CatBoost                  \\ 
                           & XGBoost                   \\ 
                           & LightGBM                  \\ \midrule
Tree Ensembles             & RandomForest              \\ 
                           & ExtraTrees                \\ \midrule
Neural Networks            & NeuralNetFastAI           \\ 
                           & NeuralNetTorch            \\ \midrule
Transformers               & FTTransformer             \\ 
                           & TabPFN                    \\ \midrule
Linear Models              & LinearModel               \\ \midrule
Instance-Based Models      & KNeighbors                \\ \bottomrule
\end{tabular}
\caption{Overview of model families and individual models.}
\label{tab:model-families}
\end{table}

\newpage

\subsection{Statistics on ground-truth labels}\label{sec:stat-gt-labels}

\begin{figure}[h]
    \centering
    \begin{subfigure}{0.45\linewidth}
        \includegraphics[width=\linewidth]{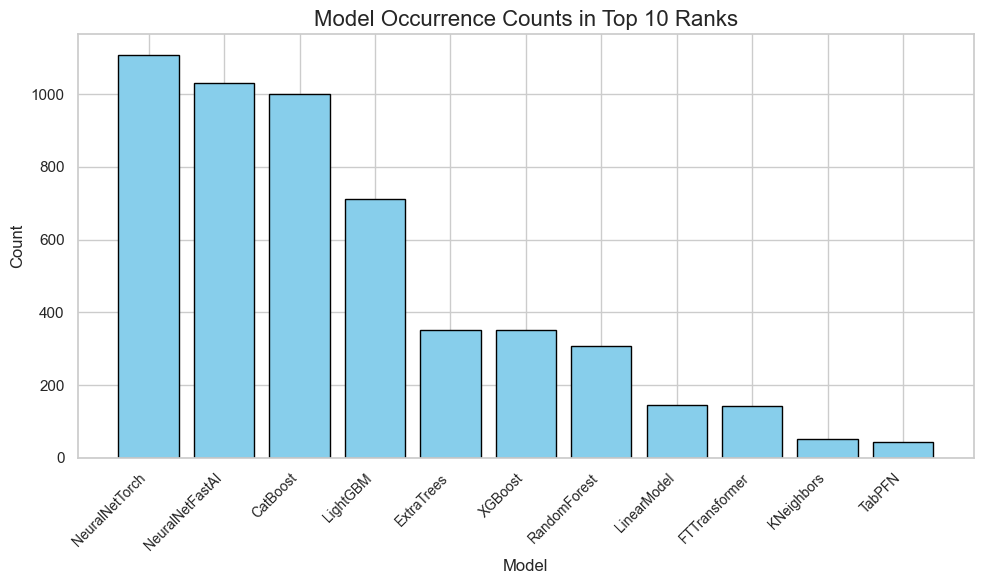}
        \caption{Top 10 model occurrences}
        \label{fig:num_samples}
    \end{subfigure}
    \hfill
    \begin{subfigure}{0.45\linewidth}
        \includegraphics[width=\linewidth]{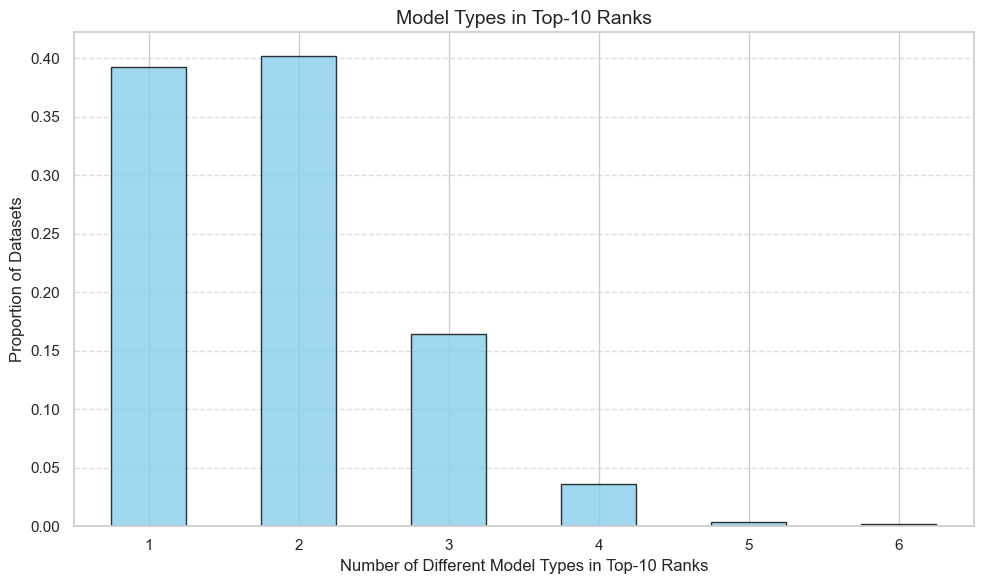}
        \caption{Top 10 model ranks}
        \label{fig:second-image}
    \end{subfigure}
    \caption{Statistic of model occurrence - AWS Portfolio}
    \label{fig:label-occurence}
\end{figure}

\begin{figure}[h]
\vspace{-0.5cm}
    \centering
    \includegraphics[width=0.95\linewidth]{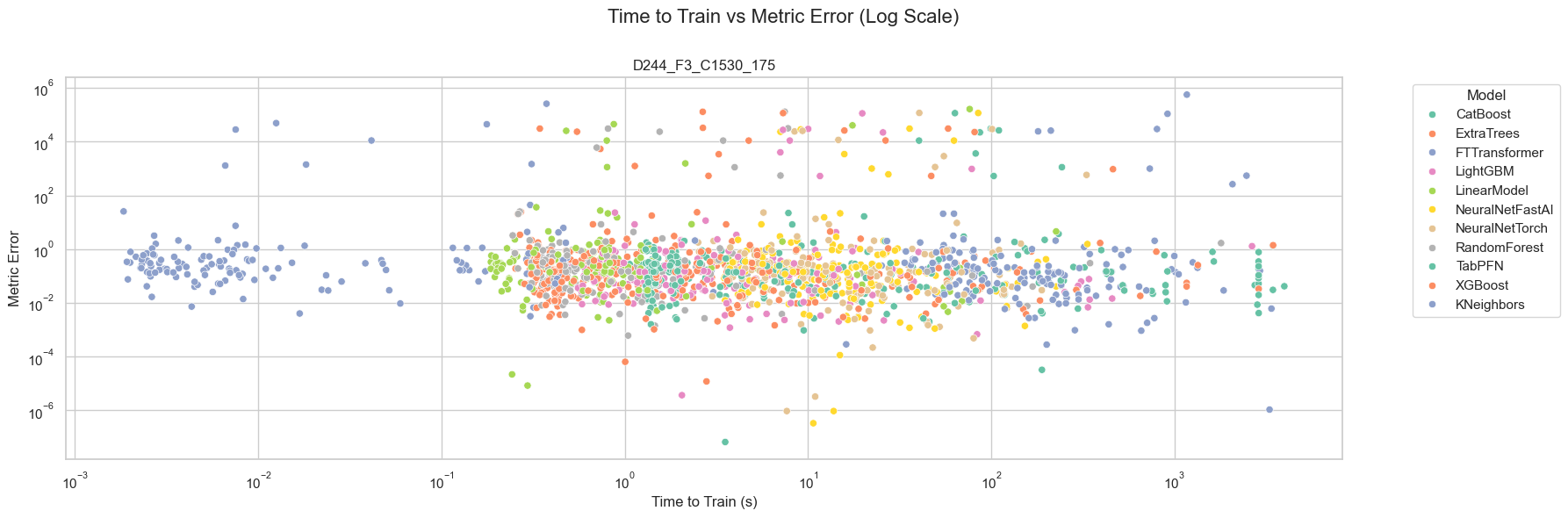}
    \caption{\centering Metric error vs. Runtime of top rank of models}
    \label{fig:stat-label-perf}
\end{figure}

For our experiments, we used the \textbf{``D244\_F3\_C1530\_175''} configuration from \cite{autogluon_tabrepo}, including 175 classification and regression datasets and 4,590 model evaluations per dataset.  Figure \ref{fig:label-occurence} presents statistics on the frequency of model types appearing in the top-10 ranking across all datasets in the portfolio. Figure \ref{fig:stat-label-perf} illustrates the relationship between metric error and runtime for top ranked model types across datasets in the portfolio. 

\vspace{-0.2cm}

\vspace{-0.2cm}

\subsection{Traditional Pre-hoc Predictors models}\label{sec:trad-pre-hp-models}

Table \ref{tab:traditional-models} presents the traditional Pre-HP we implemented to run our experiments. We employed several approaches for different types of input: a naive Euclidean distance, a K-Nearest-Neighbors (KNN), and a Random Forrest Classifier for metadata, and Cosine Similarity of TF-IDF, BERT and RoBERTa, along with a BERT Classifier for textual description input.

\begin{table}[h!]
\centering
\footnotesize
\begin{tabular}{@{}lllp{1.3cm}p{1.3cm}p{1.3cm}p{1.3cm}@{}}
\toprule
\textbf{Source} & \textbf{Method} & \textbf{Approach} \\ \midrule
\multirow{3}{*}{Meta\_data} & Euclidean Distance & Meta\_data embeddings \\
                          & KNN (n\_neigh=3) & Meta\_data embeddings  \\
                          & RFC (n\_tress=100) & Meta\_data embeddings \\ \midrule
\multirow{4}{*}{Dataset Descriptions} & Cosine Similarity & TF-IDF encoding \\
                          & Cosine Similarity & BERT encoding \\
                          & Cosine Similarity & RoBERTa encoding \\
                          & BERT Classifier & Descriptions and label \\ \bottomrule
\end{tabular}
\vspace{-0.1cm}
\caption{\centering Traditional Pre-Hoc Predictors (Pre-HP) Methods}
\label{tab:traditional-models}
\end{table}

\end{document}